\documentclass{article}
\PassOptionsToPackage{numbers, compress}{natbib}

\usepackage[preprint]{neurips_2022}

\usepackage{graphicx}
\graphicspath{figures/}
\usepackage{gensymb}
\usepackage{algorithmic}
\usepackage[frozencache=true, cachedir=minted-cache]{minted}
\usepackage{enumitem}
\usepackage{pifont}
\usepackage{color}
\usepackage{comment}
\usepackage[utf8]{inputenc} 
\usepackage[T1]{fontenc}    
\usepackage{hyperref}       
\usepackage{url}            
\usepackage{booktabs}       
\usepackage{amsfonts}       
\usepackage{nicefrac}       
\usepackage{microtype}      
\usepackage{xcolor}         

\title{Hippocampus-Inspired Cognitive Architecture (HICA) for Operant Conditioning }

\author{%
  Deokgun Park \\
  Department of Computer Science\\
  University of Texas at Arlington\\
  Arlington, TX 76019 \\
  \texttt{deokgun.park@uta.edu} \\
  \And
  Md Ashaduzzaman Rubel Mondol \\
  Department of Computer Science\\
  University of Texas at Arlington\\
  Arlington, TX 76019 \\
  \texttt{mdashaduzzaman.mondol@mavs.uta.edu} \\
  \AND
  SM Mazharul Islam \\
Department of Computer Science\\
  University of Texas at Arlington\\
  Arlington, TX 76019 \\
  \texttt{sxi7321@mavs.uta.edu} \\
  \And
  Aishwarya Pothula \\
  Department of Computer Science\\
  University of Texas at Arlington\\
  Arlington, TX 76019 \\
  \texttt{aishwarya.pothula@mavs.uta.edu} \\
}

\begin{document}

\maketitle

\begin{abstract}
Learning with many trials and errors  can be explained with Hebbian learning and implemented mechanically with model-free reinforcement learning.   However, the neural implementation of operant conditioning with few trials is not clear.  We propose a Hippocampus-Inspired Cognitive Architecture (HICA) as a neural mechanism for operant conditioning.  HICA explains a learning mechanism in which agents can learn a new behavior policy in a few trials, as mammals do in operant conditioning experiments.  HICA is composed of two different types of modules.  One is a universal learning module type that represents a cortical column in the neocortex gray matter.  The working principle is modeled as Modulated Heterarchical Prediction Memory (mHPM).  In mHPM, each module learns to predict a succeeding input vector given the sequence of the input vectors from lower layers and the context vectors from higher layers.  The prediction is fed into the lower layers as a context signal (top-down feedback signaling), and into the higher layers as an input signal (bottom-up feedforward signaling).   Rewards modulate the learning rate in those modules to memorize meaningful sequences effectively.  In mHPM, each module updates in a local and distributed way compared to conventional end-to-end learning with backpropagation of the single objective loss.  This local structure enables the heterarchical network of modules.  The second type is an innate, special-purpose module representing various organs of the brain's subcortical system.  Modules modeling organs such as the amygdala, hippocampus, and reward center are pre-programmed to enable instinctive behaviors.  The hippocampus plays the role of the simulator.  It is an autoregressive prediction model of the top-most level signal with a loop structure of memory, while cortical columns are lower layers that provide detailed information to the simulation.  The simulation becomes the basis for learning with few trials and the deliberate planning  required for operant conditioning.
\end{abstract}

\section{Introduction}

According to Thomas Reid, ``There is no greater impediment to the advancement of knowledge than the ambiguity of words~\cite{reid1850essays}.'' The  authors claim that the term ~\textit{reinforcement learning} is used to describe two different phenonmena in the AI and robotics community, namely evolution and operant conditioning~\cite{skinner1938behavior}.  Operant conditioning, also called ~\textit{instrumental conditioning}, is how a biological agent learns with trials and errors. Thorndike discovered the Law of Effect in which  pleasant outcomes associated with a behavior promote the behavior, while unpleasant outcomes discourage the behavior~\cite{thorndike1898animal}.
B.F Skinner coined the term ~\textit{operant conditioning} and called the pleasant outcomes as ~\textit{positive reinforcement}, hence   the term ~\textit{reinforcement learning} (RL) originated from it.
B. F. Skinner used the skinner box to test operant conditioning~\cite{skinner1938behavior}.   There are light bulbs and a lever in the box where a mouse is put.  If the mouse pulls the lever when the green light is on, it will get a food pellet.  However, if it pulls  when the red light is on, it will get an electric shock.  Within a few trials, it learns to pull the lever only when the green light is on.

Both evolution and operant conditioning are processes for improving behavior policies, and both can be formulated as RL problems.       More specifically, evolution is RL for episodic tasks using fitness as the reward, while operant conditioning is for continual tasks using a dopamine signal as the expected reward or the value function.
The main difference is  number of the experiences or episodes required before biological agents can change behavior policy.  Evolution requires far more experiences before agents can learn new skills, while for operant conditioning, they require only a few trials. 

In this sense, the authors claim that the AI and robotics community mastered the instinctive behavior or evolution, but they have not mastered the operant conditioning yet.  The argument as to why current AI/ML methods, especially model-free RL methods, are more similar to evolution than operant conditioning is supported by two patterns:  1) the number of training episodes are large, and 2) the trained behavior policy is brittle.

First, it is generally accepted in the AI community that the number of training episodes is one of the main challenges, referred to as the  sample efficiency issue~\cite{yu2018towards, chevalier2018babyai}.
For example, AlphaStar by DeepMind has to make many thousands of decisions sequentially where $10^{26}$  actions are possible at each time step before it gets a single reward by winning or losing a StarCraft II game~\cite{vinyals2019grandmaster}.   Yet, it learns to beat human champions using only visual observation. 
However, the core issue is that it requires much longer training - sometimes more than  200 years~\cite{alphastar}.

Second, the behavior policy developed by current AI/ML methods is known to be brittle. The visual object recognition generates erroneous classifications with adversarial attack patterns~\cite{nguyen2015deep}. In RL, the trained behavior policy becomes ineffective if the environment changes slightly. For example, if the the position of the paddle is moved a few pixels from the original setup in the Atari BreakOut game, the behavior policy learned with Deep Q-Learning (DQN) becomes ineffective~\cite{kansky2017schema}

Therefore, the authors claim that current RL replicates  evolution rather than  operant conditioning.
To tackle operant conditioning, the authors propose 1) a test framework for operant conditioning, and 2)  a cognitive architecture to pass the  test.  In the following section, we would like to clarify the problem that we think is important for the AI/ML research community.   

 \section{A Test Framework for Operant Conditioning}
 In AI/ML, the problem of developing a behavior policy in a small number of episodes is also called  few-shot learning~\cite{wang2020generalizing} of which  transfer learning has been one of approaches for~\cite{pan2009survey}.  In transfer learning, the lower layers are pre-trained and a few top-most layers are fine-tuned with a new dataset that  reduces the number of labelled datasets.   This also includes using pre-trained embeddings in natural language processing~\cite{brown2020language}.  Another major approach is meta-learning~\cite{hochreiter2001learning,finn2017model} where a meta-learner uses knowledge across multiple tasks.
 For example, the meta-learner might use a performance among multiple tasks to provide an initial parameter for the new tasks that are sensitive or quickly reducing the objective function~\cite{finn2017model}.    An additional approach is learning to predict whether two images belong to the same class or different classes using the Siamese network~\cite{koch2015siamese}.  In this approach, the training data is built by sampling positive and negative pairs where a positive pair means both samples belong to the same class and vice versa.

 Rather than discussing the limitations of individual approaches or models, the main limitation of prior works is that the test scenario or application is not sufficient or ecologically valid for solving operant conditioning.  Many models are tested for the image recognition or classification regime.  While the few-shot learning capability in the image recognition is a necessary component, it is hard to tell that it is  sufficient for the validation of operant conditioning that requires perception and action (PAC).  Even in robotics, where PAC is required, most prior researches are for learning by imitation or demonstration`\cite{duan2017one, dance2021conditioned}.   Therefore, a new test in the embodied agent  that are related to operant conditiong is required.  

\begin{figure*}[htb]
  \includegraphics[width=0.8\textwidth]{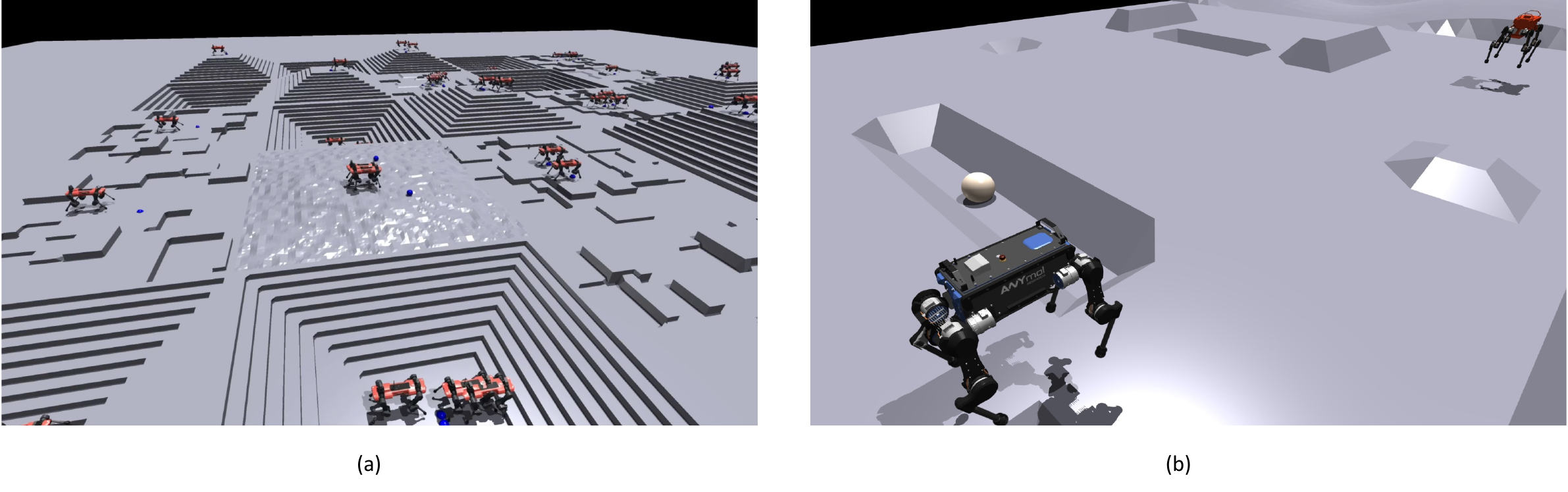}
  \centering
  \caption{The screenshot for the proposed test framework.  (a) In Phase 1, the agents are trained using traditional reinforcement learning. In this example, four-legged robots are trained to pick up blue balls and avoid predators.  This phase may use GPU accelerated parallel training to reduce training time.  (b) In Phase 2, the agent encouters novel elements, such as new types of reward objects/predators and novel mechanisms resembling skinner box. }
  \label{fig:screenshot}
\end{figure*}

 To test operant conditioning, we propose the ~\textbf{Virtual Skinner Box} test framework.  The test is composed of two phases.  In the first phase, we train an embodied agent in a simulated environment to have basic behavior policies as mice have.  For example, it can be trained to pick up blue balls to get energy or to avoid predators.  In the second phase, we introduce new elements to the simulated environment such as new type of food items, predators or novel mechanisms resembling skinner box.  The environment might have a lever and a light, where pulling a lever when the light is off results in a negative reward and vice versa.  We can say that the agent has an operant conditioning capability when the agent learns new behavior policies in a few trials say less than ten trials.  Figure~\ref{fig:screenshot} shows the test framework with the legged robots as an agent.

 \section{A Cognitive Architecture for Operant Conditioning}

 To solve the Virtual Skinner Box test, authors propose a cognitive architecture that is motivated by the brain.  (Please note that the following biological and neuroscience explanation  is to serve the design of the next generation computational architecture.  The goal is not biological correctness, but deriving essential design principles.  Therefore, the explanation is brutally simplifying  complex biological phenomena to serve this goal.)
 A brain is to generate appropriate behaviors.  Biological agents that do not move, do not have brains.  Trees don't have brains.  Sea squirts have brains (neural cells) when they are moving in the ocean, but after settlement they don't move and eat (resorb) their brins~\cite{cosmides1997evolutionary}.
 Some behaviors are innate (instinct), and others are acquired (learned)~\cite{tinbergen1951study}.  Therefore, components in the brain may be divided into two parts; one part participates in instinctive behaviors and the other in learned behaviors.  Instinct is the prevalent  mechanism for fishes, amphibians, and reptiles.  Learned behaviors in the form of the operant conditioning   appeared in  mammals and rodents.  Interestingly, the neocortex first appeared in mammals and rodents.  Therefore, it is possible to attribute the operant conditioning with the neocortex.  And the rest of the brains or the subcortical structure is responsible for the instinct.    The  neocortex is also composed of gray matter and white matter.  Gray matter forms a six-layered thin sheet crumpled in the skull and white matter comprises the fiber bundle (neural tract) for the long range connection  between different area of gray matter  or between gray matter and subcortical structures.

\begin{figure*}[tb] \includegraphics[width=0.7\textwidth]{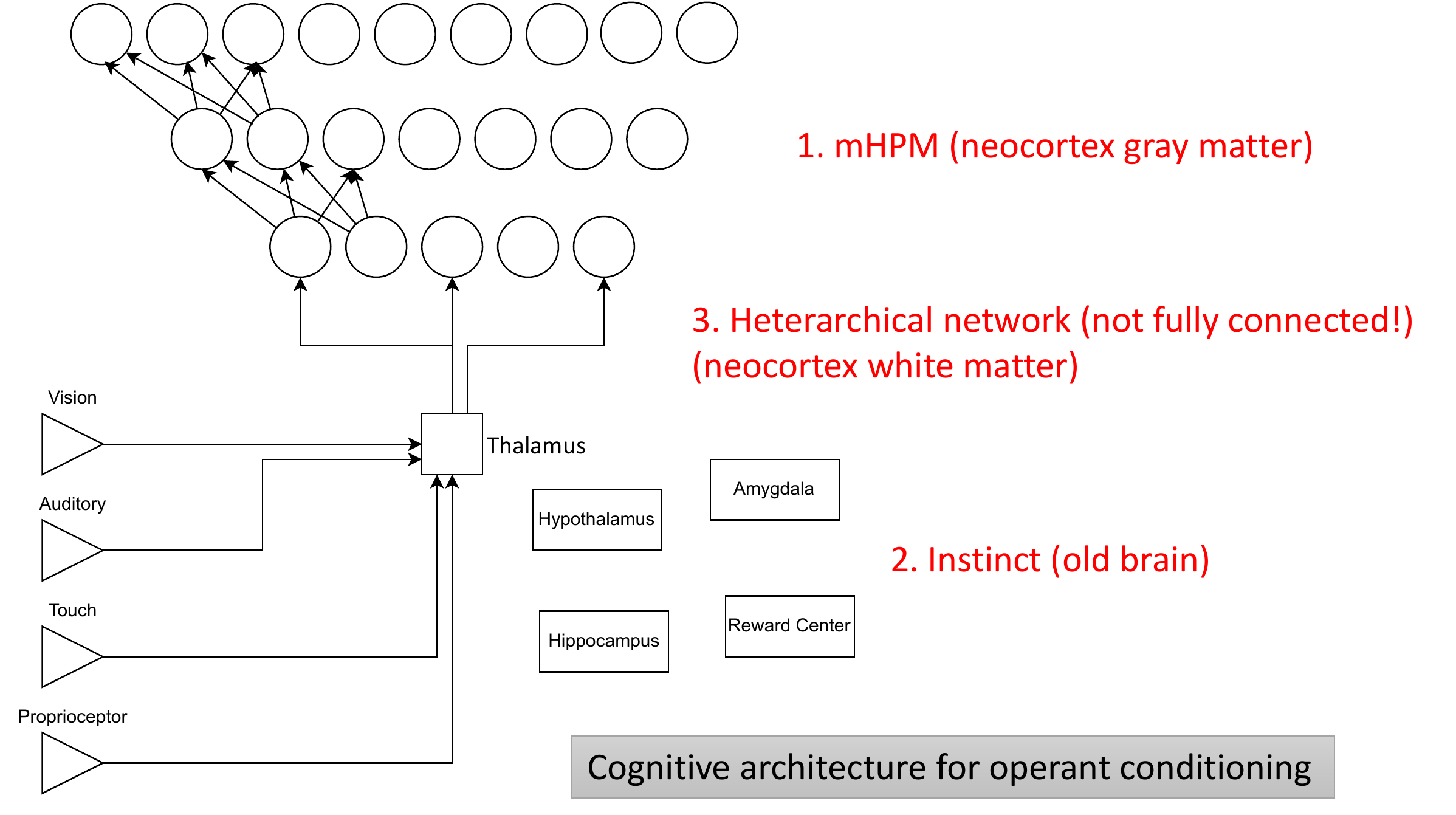}
  \centering
  \caption{The proposed cognitive architecture is composed of three components. 1. modulated Heterarchical Prediction Memory (mHPM) models the cortical column or the canoical microcircuit of the neocortex.  2. There are a few innate modules modeling various subcortical structures. 3. These mHPM modules and innate modules are connected with heterarchical connections. }
  \label{fig:cognitive_architecture}
\end{figure*}

 Inspired by this, the proposed cognitive architecture is composed of three components:
 
\begin{enumerate}
\item A modulated Heterarchical Prediction Memory (mHPM) modeling the neocortex gray matter,
\item A set of innate modules modeling subcortical structures such as thalamus, amygdala, hippocampus, and the reward center, 
\item And a heterarchical network between mHPM modules and innate modules. 
\end{enumerate}
Figure~\ref{fig:cognitive_architecture} shows the diagram for the cognitive architecture.  
mHPM modules are learned or updated as current deep learning approaches, while innate modules are manually programmed by researchers and fixed during the simulation.

\subsection{Modulated Heterarchical Prediction Memory (mHPM)}

Different areas of the neocortical gray matter serve different functions such as motor, vision, auditory, language, and so on.  Many conjecture these functions are different because  the input and output neural pathway connecting each area is different, but the blueprint for different areas of the neocortex is  same~\cite{mountcastle1997columnar, buxhoeveden2002minicolumn}.  For example, if a specific area is connected with the auditory and vocal tract, the area will serve the function related to the language.  However, if  other area is connected with tactile sensors and proprioceptors, it might serve motor-related functions.  A basic unit for modular functionality is called a cortical column~\cite{mountcastle1997columnar} or a canonical microcircuit~\cite{bastos2012canonical}.  This hypothesis is supported by two arguments.  First, when a specific area is nonfunctional due to lesions, the function associated with that area might be restored in an other area by brain plasticity~\cite{eagleman2020livewired}.  If the blueprints  of the cortical columns are all different, this would not be possible.  Second, if the blueprint of a cortical column is different for a specific function,  all parameters or weights of the neocortex would  have to be  written in the genetic code.  Then, animals  should have different size of the genetic code proportional to the size of neocortex.  But the size of the genetic code between humans, rats, and reptiles are not showing this pattern.   It is anticipated that the code for the cortical column  will be replicated many times in a modular way to construct neocortex, while the codes for other subcortical structures  will be replicated only one time to build  functional innate organisms.  If we assume the same blueprint hypothesis for the cortical columns or canoical microcircuits, important questions are what function each cortical column does and how we can model the mechanism of a cortical column.

Our  hypothesis is that the working principle of cortical columns  can be explained as  \textbf{modulated heterarchical prediction memory (mHPM)}.
Many conjecture that the main function of a cortical column is to predict the next vector signal given the sequence of vectors.  It is also called an  autoregressive (AR) model~\cite{dai2015semi,  friston2009predictive, rao1999predictive}. 
Recent advances in language model and image generation with self-supervised learning or semi-supervised learning are based on this~\cite{brown2020language, chen2020generative}.  

Perception is a prediction in the sensory input vector sequence; action is a prediction in the motor command vector sequence. 
Consciousness is a prediction of the vector sequence in the top layer.  
For most verbal thinkers, it is implemented as the prediction in the articulatory rehearsal component (articulatory loop) in the Broca's area commonly referred as~\textit{inner voice}~\cite{buchsbaum2013role, muller2006functional}.
For most verbal thinkers, it is implemented as the prediction in the articulatory rehearsal component (articulatory loop) in the Broca's area commonly referred as~\textit{inner voice}~\cite{buchsbaum2013role}.

\begin{figure}[bt]
    \centering
    \includegraphics[width=0.7\textwidth]{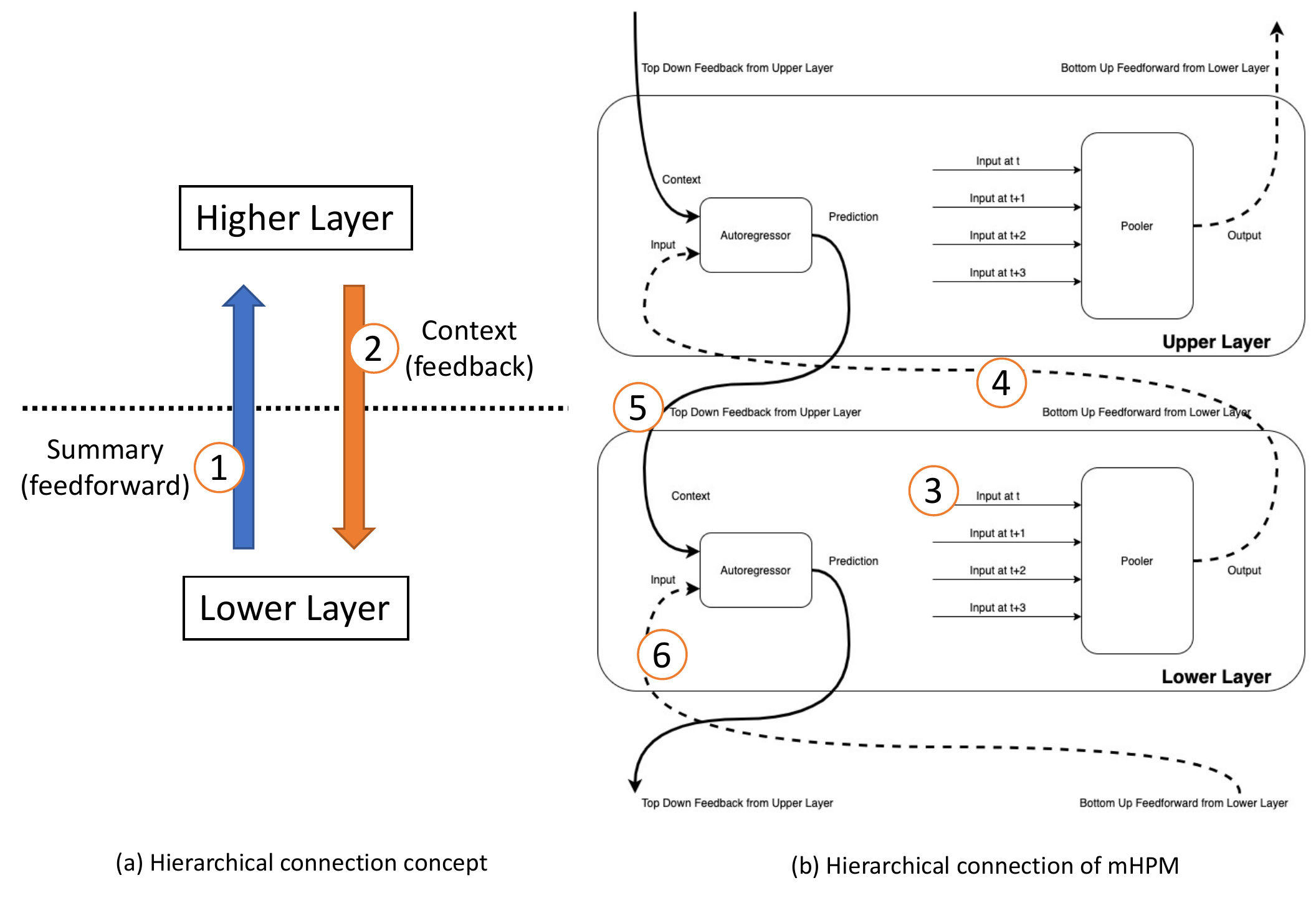}
    \caption{Workings of two layers in Heterarchical Prediction Memory (HPM).  Lower layer feeds \ding{192} a summary as an input signal for higher layer that is a feedforward signal. Higher layer sends \ding{193} a context for the lower layer as a feedback signal. \ding{194} $k$ inputs from the lower layer are summarized with pooler which is an autoencoder. \ding{195} A summary from the lower layer are feedforward as an input for the autoregressor in the higher layer. Autoregressor predicts \ding{196} next input with the summary input from the lower layer and context signal from the higher layer. \ding{196} This prediction is feedbacked to the lower layer and used as a context for autoregressor in the lower layer.  }
    \label{fig:mHPM}
\end{figure}

One fundamental issue of the AR model is dealing with long-term dependencies. Hierarchical structures have been used to solve long-term dependencies, such as hierarchical RNN~\cite{chung2016hierarchical, du2015hierarchical, qiu2019neurally} or hierarchical RL~\cite{botvinick2012hierarchical, barto2003recent}. We propose to improve them by two ideas, 1) providing feedback signal from higher layer as context, and 2) doing online and local updates instead of the end-to-end learning approach. In mHPM, each layer has an AR module which predicts a next vector given input from the lower layer. The prediction is routed to the lower layer as a feedback signal, which provides a context for the AR module in the lower layer.  The lower layer provides a feedforward signal to higher layer that is the summary of $k$ input sequence in the lower layer.  Autoencoder (AE) model is used to make a summary of $k$ inputs. Figure~\ref{fig:mHPM} represents this structure when $k = 4$ and Listing 1 shows the  pseudo code. Please note that the lower layer has to run $k$ prediction steps to generate one input for the higher layer. Therefore, temporal frequency and the convergence speed are slower in higher layer which is supported by the biological evidence ~\cite{kiebel2008hierarchy, murray2014hierarchy, runyan2017distinct}.

    
  


\begin{figure}[bt]
    \centering
    \includegraphics[width=0.8\textwidth]{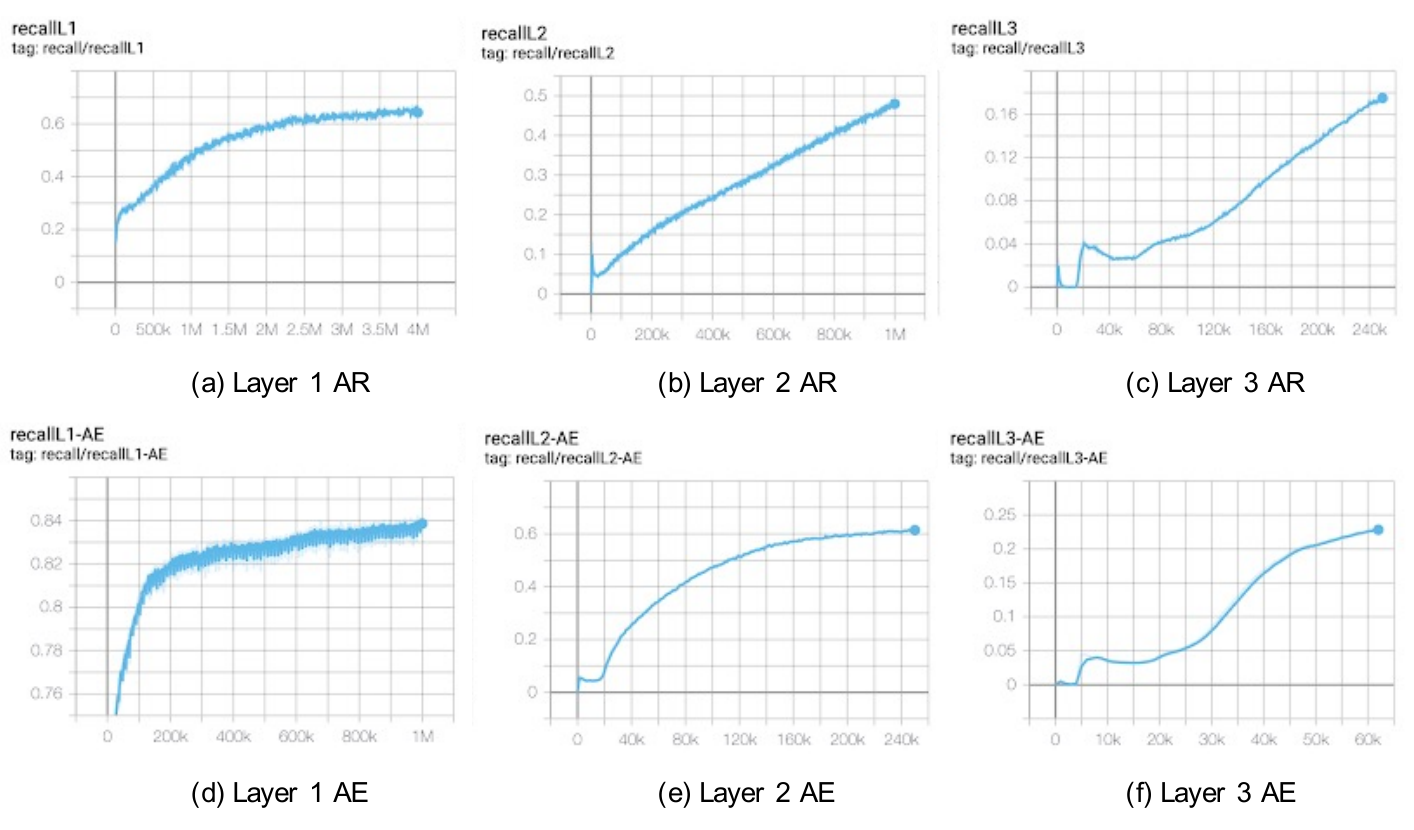}
    \caption{Performance of mHPM for character-level language modeling task with corpus size of 0.5M characters. The chart shows a prediction accuracy for autoregressive (AR) and autoencoder (AE) for layer 1, 2, and 3. (the higher the more accurate). It is interesting that the prediction accuracy of higher layers do increase considering that their input sequence is from the embeddings of an AE model in a lower layer which is also learning from scratch. }
    \label{fig:performance}
\end{figure}

\begin{figure}
    \centering \includegraphics[width=1\textwidth]{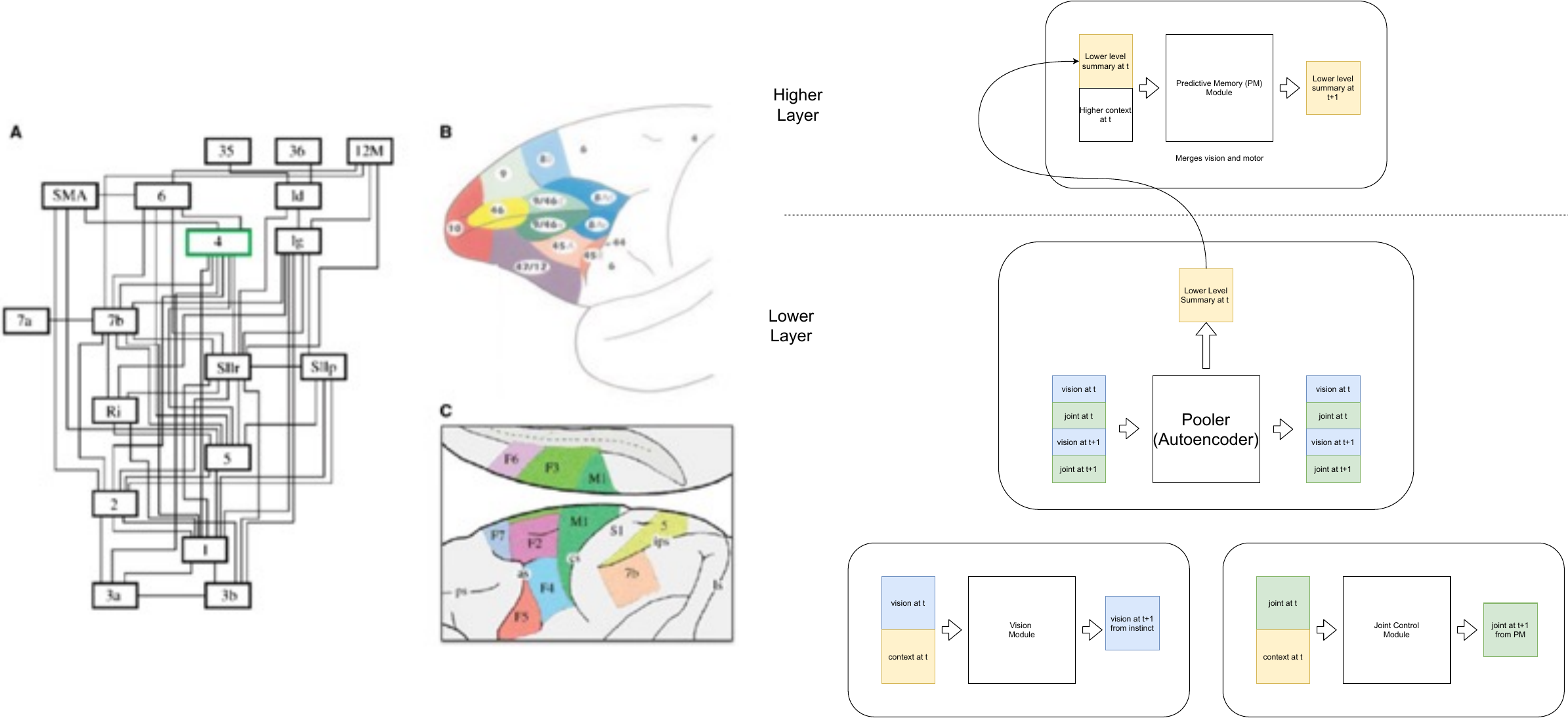}
       \caption{Left: The somato-motor hierarchy for a monkey brain by ~\cite{felleman1991distributed}. Right: Merging Visual pathway and motor pathway by merging signals using autoencoder and feed it forward as input signal for  higher layer. Prediction from the higher layer is feedbacked to the both lower layers (sensory and motor) as the context input. }
       \label{fig:monkey_brain}
\end{figure}

Figure ~\ref{fig:performance} shows the convergence when we used three layers. This was trained as a character-level language model with the training corpus size of 0.5M characters. This shows the preliminary evidence that this structure converges for sequence modeling. However, the previous end-to-end learning models such as attention or recurrent models outperform in language modeling. However, the main benefit of the proposed structure is that the AR and AE model in each layer can be updated online and locally as explained in figure~\ref{fig:cognitive_architecture} and in Listing 1.
In end-to-end learning models, the ground-truth for prediction is available only in the bottom layer. Therefore all higher layers are updated with the backpropagation of prediction error of the bottom layer.  In mHPM, all AR and AE models have the local ground-truth in every time step. In this sense, every module in mHPM is self-sufficient for self-supervised learning. This enables the parallel and asynchronous update of individual modules. Therefore, we can build a complex heterarchical network. Figure~\ref{fig:monkey_brain}  shows the somato-motor hierarchy for a monkey brain that is an example of a heterarchical network  ~\cite{felleman1991distributed} (left) and how visual and motor pathways can be combined in mHPM (right). 

\begin{listing}[bt]
\begin{minted}{python}
class Layer:
    def __init__(self):
        self.lower = None   # Lower layer
        self.AR = AR()      # Autoregressive model
        self.AE = AE()      # Autoencoder model
        self.pred, self.prevInput = rand(n), rand(n)
    def feed(self, context):
        buffer = []
        for i in range(k):
            self.pred = self.AR.predict(self.prevInput, context)
            input = self.lower.feed(self.pred)
            error = self.pred - input
            AR.update(error)
            buffer.append(input)
        summary = AE.feed(buffer)
        return summary
\end{minted}
\caption{Pseudocode for the flow of the Layer. Method \textbf{feed} gets the summary from the lower layer with the previous prediction as the context. Then, after receiving k inputs, it generates a summary with autoencoder and return it as the input for higher layer.  }
\label{listing:3}
\end{listing}





 \begin{figure}
  \centering  \includegraphics[width=0.4\textwidth]{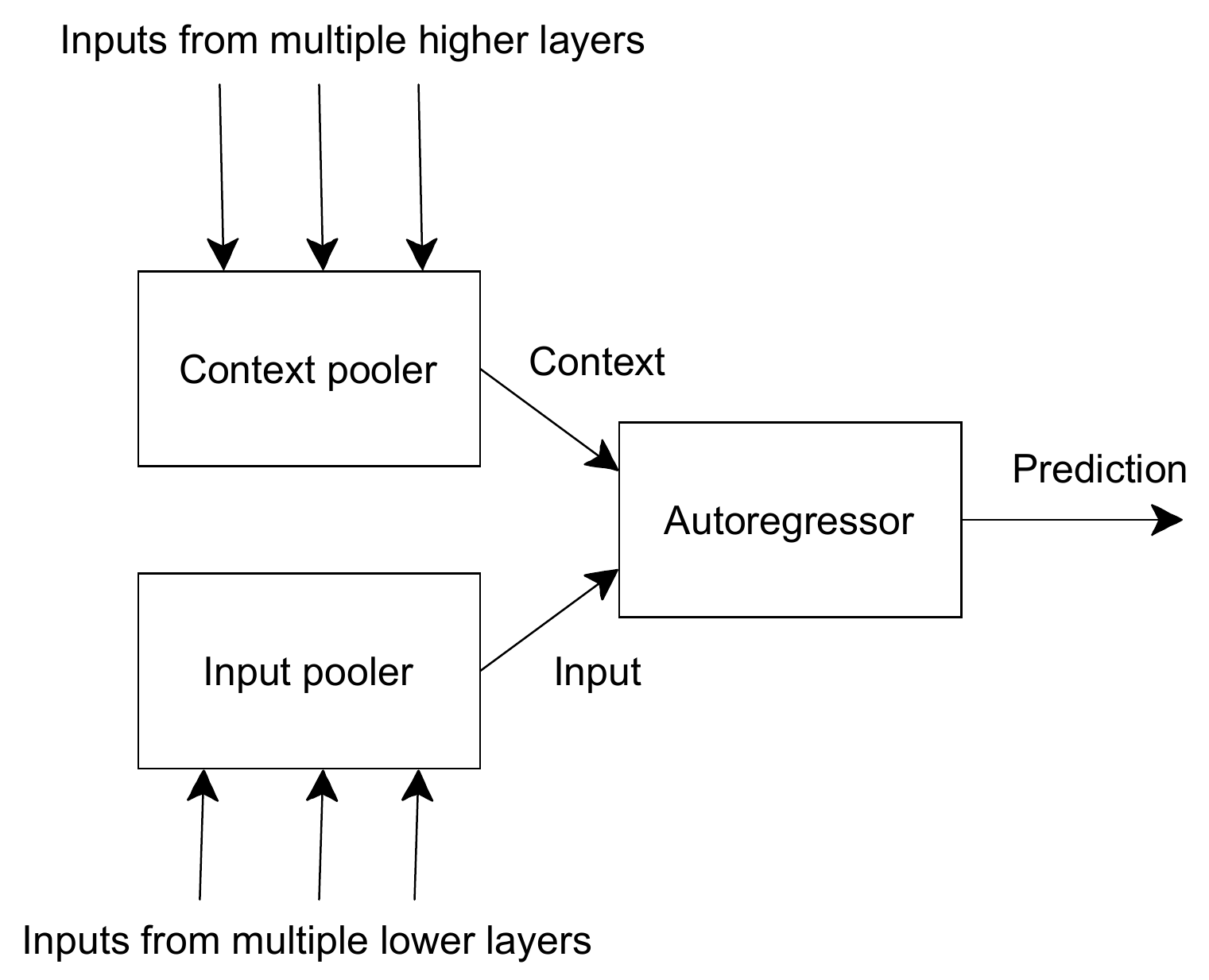}
  \caption{The mHPM module for a hetearchical network}
  \label{fig:instinct}
\end{figure}
A heterarchical network is a graph where two nodes connected by a link contain an additional information about which node is higher. Compared to a hierarchical network which is  represented as tree structure, a node in a heterarchical network can have multiple parents node. For example, reaching a hand to grasp an object requires multiple pathways: the what and where pathway in vision cortex, motor cortex, and somatosensory cortex work together.
 In mHPM, multi-modal signals such as motor, vision, auditory can be combined with AE to build a summary of inputs.  Therefore, each AR model will have inputs from multiple lower layers combined in AE.  Similarly, the context signals from multiple higher layers are also combined in AE to build a summary context.  In this way, complex heterarchical network such as Figure~\ref{fig:monkey_brain} (left) can be modeled.

However, this  mechanism  simply memorizes its sequence regardless of importance or correctness.  In the eyes of a newborn baby, everything will be new.  There are important sensory inputs, such as the face of the caregiver, and meaningless ones, such as white noise on TV screen.  Similarly, a baby moves its arm randomly as motor babbling.  But some motor command sequences yield meaningful behavior such as grasping toys, while others do not.  Therefore, we need a mechanism for learning only meaningful sensory and action sequences.  In mHPM, the update rate of AE and AR models are modulated by a reward signal to memorize only meaningful sequences.  When there is a reward, the update rate will increase and the sequence will be more likely memorized.   
The effect of this change is global and lasts for a while (about 2.5 seconds in the human).  This modulation based memorization can explain why rats can learn quickly when there are large rewards either positive or negative, and why we can learn procedual memory if we repeat a certain behavior repeatedly without any rewards such as riding a bicycle or writing the mirrored alphabet.

\begin{figure}[bt]
    \centering
    \includegraphics[width=0.8\textwidth]{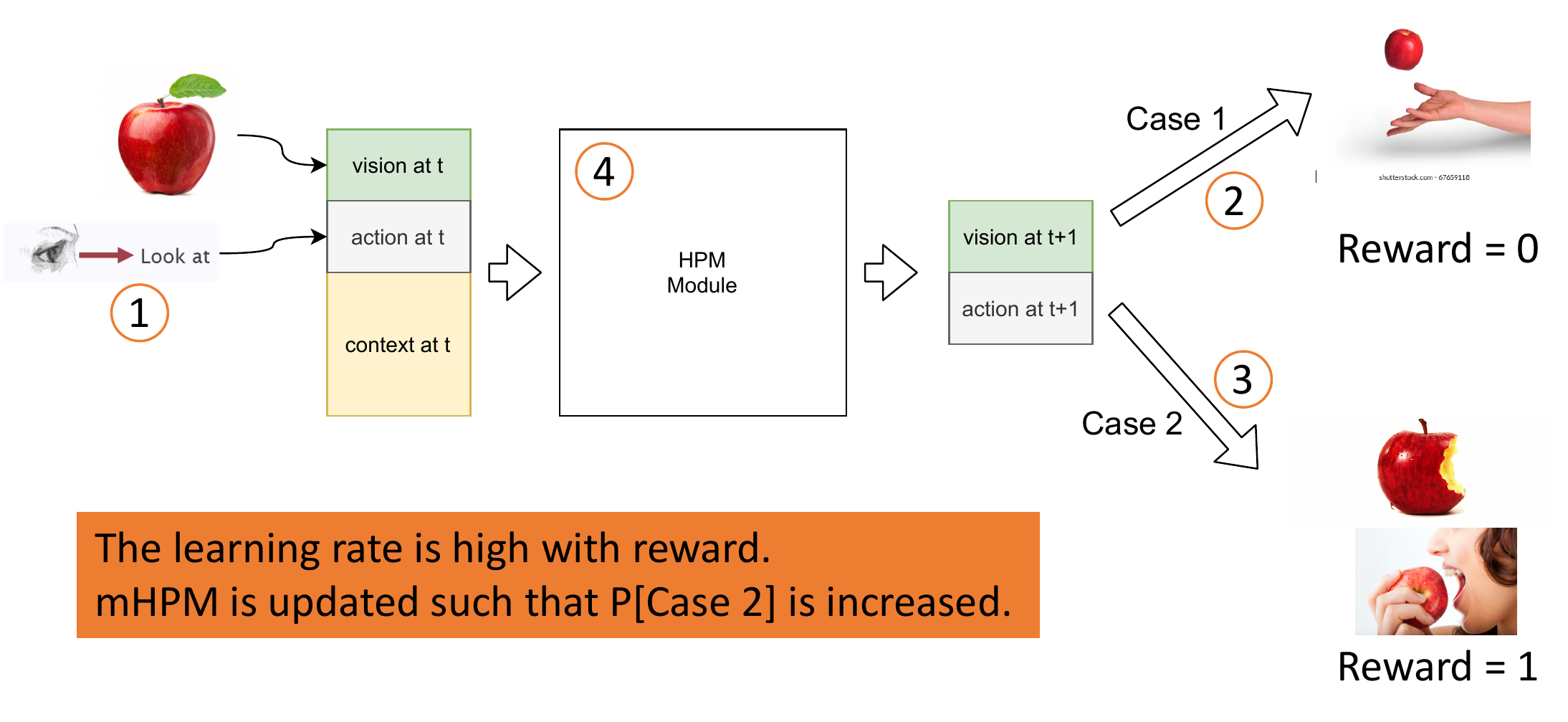}
    \caption{This explains how reward modulates the learning. Given \ding{192} a state of looking at an apple, \ding{195} the mHPM may decide whether \ding{193} to throw it or \ding{194} to taste it. \ding{194} Eating generates a reward while throwing does not. When there is a reward, the learning rate of mHPM is high and vice versa. Because the learning rate of \ding{194} eating case is high, mHPM is more likely to predict tasting action than throwing action.   }
    \label{fig:mHPM}
\end{figure}


\subsection{Engineering Specialized Modules for Instincts}

Considering the prominent role of the neocortex, it is easy to regard it as a main role, and specialized parts as optional supporting roles.  But those  parts are the original brain for reptiles.  The neocortex later joined as an auxiliary supporting unit for mammals.  An analogy in computer architecture would be the CPU and RAM.  Historical computers from mechanical calculators to ENIAC did not have a RAM to hold a program and data.
ENIAC had accumulators that held about 20 numbers and used plugboard wiring as a programming method that took weeks to configure.  Still, ENIAC was a general-purpose computer and Turing complete.  Modern computers using RAM for program memory in von Neuman architecture came later as an additional feature to an already general-purpose computer. 
However, the modern computer has a vast RAM, and that it is a key contributor to its flexible and universal functionality.  The neocortex is like RAM.  Both are large scale and uniform, and came later as a novel feature for an already-fully-functioning general purpose information processing system.

Instinct is a shortcut that enables a reasonable behavior policy in the life time of individual biological agents. Given infinite time, an agent with  learning capability might learn all the things that an intelligent animals can do without the help of instinct.  But in the biological agents, most behaviors of  biological agents are not based on learning but instincts.   Let us consider a rabbit that has never seen a wolf before.  If the rabbit tries to learn the appropriate behavior by trial and errors when it does encounter a wolf for the first time, it is too late to update its behavior policy based on the outcome of random exploration. The rabbit should rely on the instinct instead. Natural environments are too hostile to use learning as the primary method of building a behavior policy.  Therefore, we need to study how we can incorporate instincts in the study of AI.  The current SOTA tends to be more homogeneous in its structure emphasizing learning only.



There are different implementation mechanisms for instinct including reflex, neurotransmitter-based modulation, and special-purpose structures. For example, raising arms when tripping, sucking, crawling, and walking are examples of reflex. Reflex relies on dedicated neural circuits. It is useful when it is okay that the response is rigid or fixed  and the reaction duration is instantaneous.  However, when a rabbit hears a wolf cry, the reaction needs to be flexible depending on the context. The reaction state should be maintained over longer time span.  Modulation using neurotransmitters or hormones is effective in those cases, because its effect is global, meaning various areas of brain can respond according to it. And it lasts a longer time span before it is inactivated. Finally, the hippocampus or basal ganglia are special-purpose structures that solve  particular problems such as memory consolidation or decisions among conflicting behavior plans~\cite{brown2004laminar}.


Let's call the minimum set of instincts for operant conditioning as ~\textit{knowledge instinct}.  Below are components that we conjecture essential: 1) minimum reflexes, 2) basal ganglia, 3) amygdala, 4) reward system, 5) hippocampus.

\paragraph{Minimum reflexes}  In terms of legged robots, this reflexes include walking, picking up foods, avoiding predators and so on.  As explained earlier, current RL methods can be used to build reflexes effectively.  Hiearchcial structure emerges.  For example, a lower level module might drive  behaviors such as controling each joint to walk, while a higher level module drives  the walking direction based on the visual input~\cite{merel2018hierarchical}.

\begin{figure}
  \centering  \includegraphics[width=0.35\textwidth]{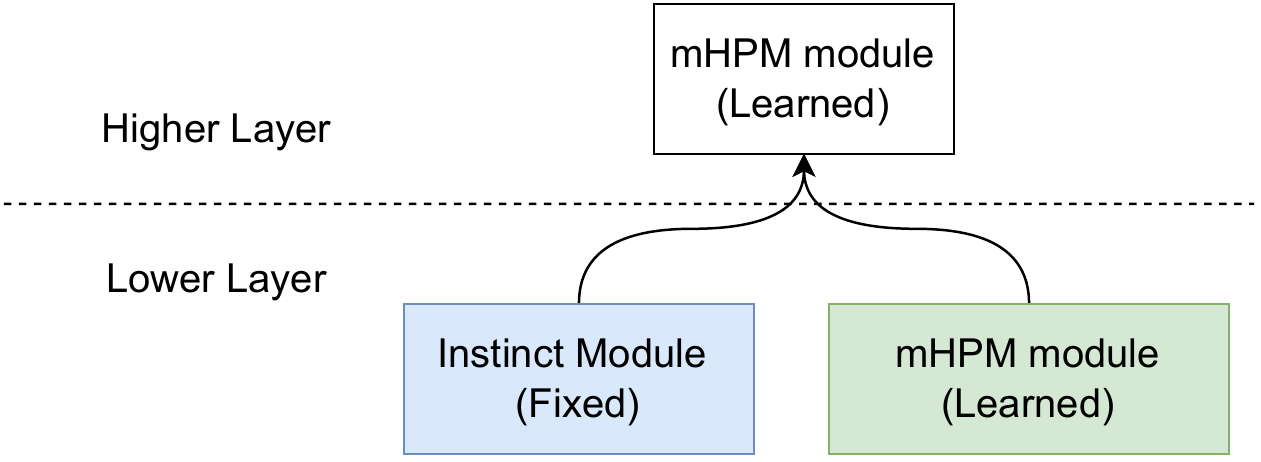}
  \caption{Merging the innate and learned behavior}
  \label{fig:instinct}
\end{figure}
\paragraph{Basal ganglia} ~\textit{Basal ganglia} resolve conflicts involving multiple behavior options ~\cite{brown2004laminar}.
For example, primates have a reflex that foveates to a moving object (pro-saccade).
However, monkeys can be trained to foveate to moving target if the fixation point is red, and to move their eyes in the opposite direction of the target if the fixation point is green (anti-saccade) ~\cite{everling1999role}. 
In mHPM, artificial basal ganglia is implemented such that the lower layer is composed of an innate behavior module and a learning module.  The lower-layer learning module might learn a new behavior.  Higher layer  module moderates the instinctive behavior and the learned behavior.

\paragraph{Amygdala}  ~\textit{Amygdala} controls the operation mode of brain among 1) fight or flight, 2) automatic operation also called type I or system I~\cite{kahneman2011thinking} (such as driving or riding bicycle), 3) focused deliberate operation also called type II or system II, and 4) idle or boring mode.  The mechanism of amygdala can be implemented as a simple classifier of current sensory signals.

\paragraph{Rewards  system} A  reward system can be implemented as the classifier with a few internal sensory signals for hunger or thirst.  A dedicated mHPM module  connected to reward system will predict the outcome of reward system.   This replicates the prefrontal cortex.

\paragraph{Hippocampus} Hippocampus is related to the consolidation of memory.  It plays an important role in the proposed cognitive architecture that will be explained in the next section.

\subsection{The Innate Hippocampal Program}
What might be the evolutionary driving force that motivated operant conditioning?
Instincts were good enough that reptile and fish prevailed.
We conjecture that the foraging task provided the evolutionary drive towards operant conditioning.  Imagine a dark cave.  Foods, water, and entrances are scattered.  With a few explorations, an animal needs to learn to effectively visit places in a planned sequence.  This is also called as ~\textit{path integration capability}, and many animals this problem solve differently including bees, desert ants, migratory fish, and mammals.  However,   the solution developed by mammals was extensible to other problems such as episodic memory, planning, and operant conditioning.

\begin{figure}[bt]
  \centering  \includegraphics[width=0.84\textwidth]{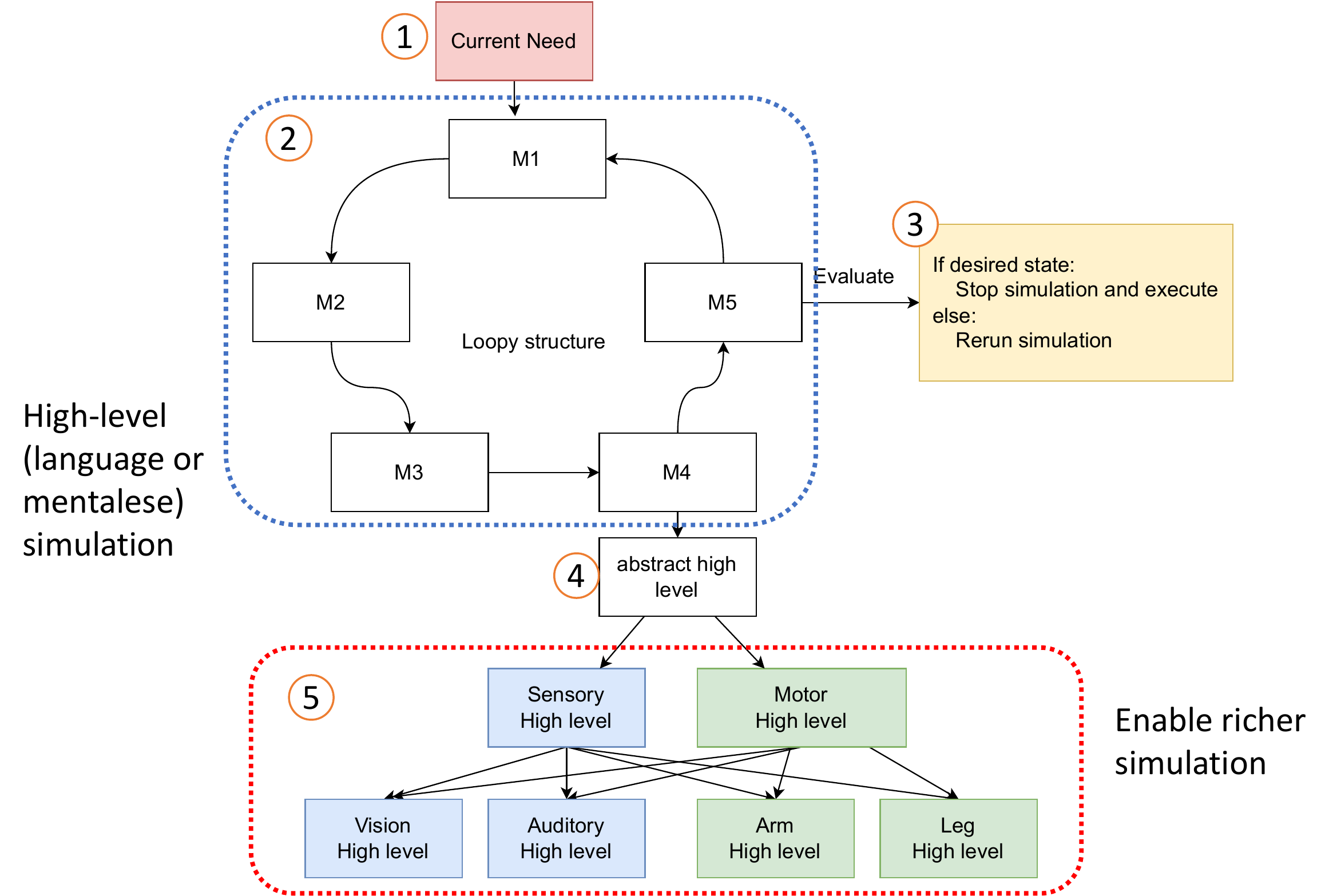}
  \caption{Hippocampal simulation(preplay) is a mechanism implementing planning, navigation, and type II behavior. 
Let's assume that there is a need that can not be met in automatic operation mode.  Artificial amygdala changes the brain mode to focused deliberate mode by  disabling motor execution, and putting the current need vector (\ding{192}) to the first slot (M1) of the loop memory in artificial hippocampus (\ding{193}). The hippocampus loop structure predicts the next situation in high-level.  For example, if the need is hunger, the next prediction might be to go to a refridgerator and the next prediction is to open the refridgerator and so on.  Please note that the prediction is the top-level of mHPM (\ding{195}) and it activates lower layers that provide the richer multi-modal simulation with encoded world models (\ding{196}).  For example, the lower-layers might provide visual image that is predicted on opening the refridgerator including its contents.  The simulation continues to evaluation step (\ding{194}) where it is determined if desired state is achieved.  If it is achieved, the motor execution is enabled and the contents of loopy structure provide high-level context or working memory for plan execution.  If not, the simulation is reset with current need and continues. }
  \label{fig:hippocampus}
\end{figure}

The solution developed by mammals was to build a cognitive map of the exploration space using the hippocampus and related memory networks~\cite{suddendorf2006foresight, mullally2014memory, george2021clone}. The hippocampus can be modeled as loop structured memory that enables repetition of a sequence to write down to cortical columns for episodic memory~\cite{olafsdottir2018role, replay}.  This structure is required because cortical columns and mHPM modeling them require repetition of a sequence to memorize it.  But the event happens only one time and the sequence of foraging exploration is also rare.  The hippocampus loop structure allows the repetition of previous events (replay) and the simulation of future events (preplay)~\cite{dragoi2011preplay}.  Replay and preplay are supported by world knowledge encoded in neocortex gray matter.  In the proposed cognitive architecture, the loop memory of the hippocampus is connected to top-most layers in mHPM.  The predictions in the higher layer is used as context for lower-layers and activate the whole network in top-down ways. Lower-layers provide detailed sensory activation thus enabling richer simulation.  This simulation provides the mechanism for operant conditioning.  Ha and Schmidhuber showed that world models can be encoded in  autoregressive models~\cite{ha2018world}.  This proposal builds an innate hippocampal program upon it that is explained in the Figure~\ref{fig:hippocampus}.
The mHPM module that is closely related to the reward system is used to evaluate the value of current prediction during the simulation.  This mHPM module  implements the role of prefrontal cortex in the human brain.

This hippocampal program explains how path integration, cognitive maps, planning, type II (system II) thinking, and operant conditioning are implemented.


The central question is how to build a mechanical model of a rodent brain with the operant conditioning and maze navigation capabilities.      This architecture might be used to design controllers for robots with a learning capability of dogs in artificial intelligence (AI) community.   Furthermore, this architecture might provide a framework for understanding  mammal brains in neuroscience community.
Our architecture is based on how Kahneman distinguished fast System 1 thinking and slow System 2 thinking ~\cite{kahneman2011thinking}.  Fast System 1 thinking can be explained with Hebbian learning, and implemented reasonably well in current artificial neural networks (ANN).  However, it is still challenging to build System 2 thinking.  We propose a hypothetical cognitive architecture that is inspired by the interplay between hippocampus and neocortex.

\begin{figure}[tb]
  \includegraphics[width=1\linewidth]{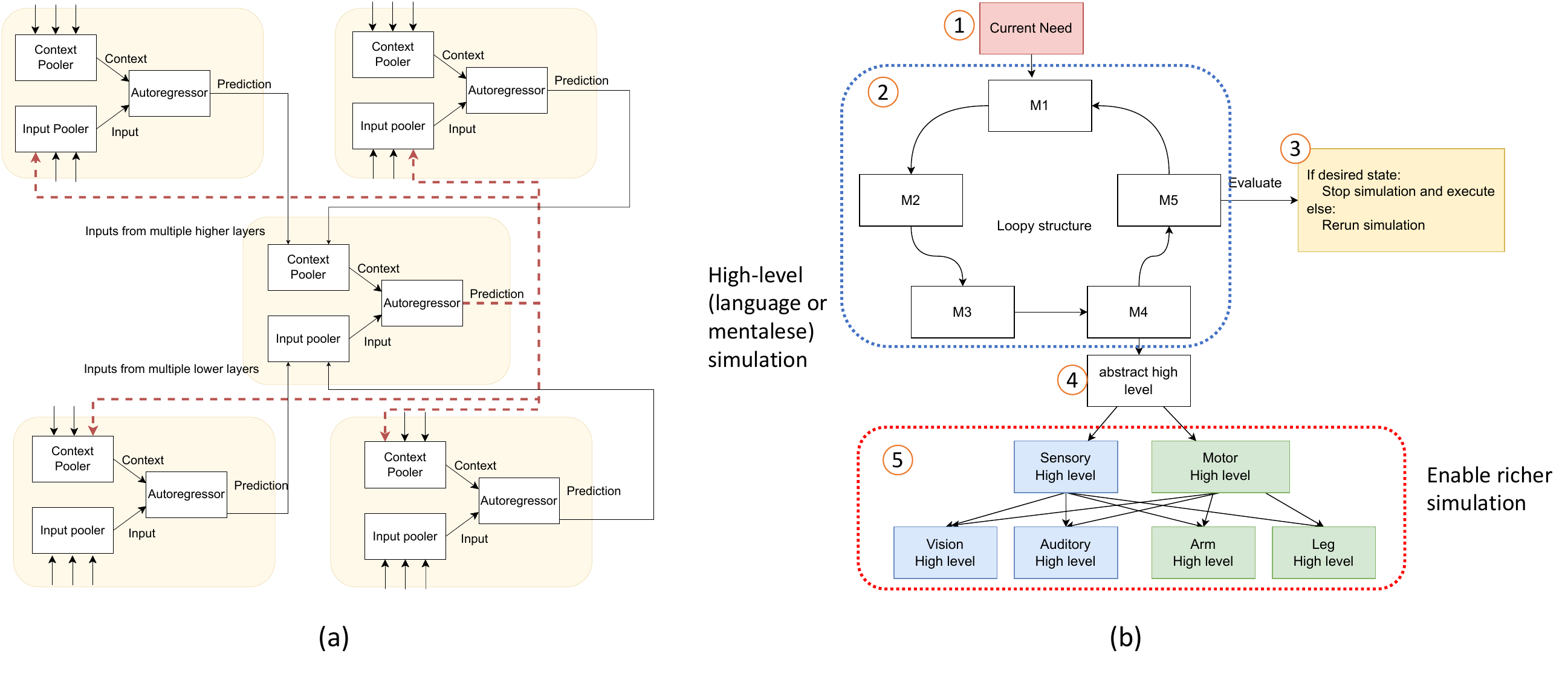}
  \centering
  \caption{(a) mHPM models the cortical columns as networks of memory  (b) Simulation in HICA.    
Let's assume that there is a need that can not be met in automatic operation mode.  The artificial amygdala changes the brain mode to the focused, deliberate mode by  disabling motor execution, and putting the current need vector (\ding{192}) to the first slot (M1) of the loop memory in artificial hippocampus (\ding{193}). The hippocampus loop structure predicts the next situation in high-level.  For example, if the need is hunger, the next prediction might be to go to a refridgerator and the next prediction is to open the refridgerator and so on.  Please note that the prediction is the top-level of mHPM (\ding{195}) and it activates lower layers that provide the richer multi-modal simulation with encoded world models (\ding{196}).  For example, the lower-layers might provide a visual image that is predicted on opening the refridgerator, including its contents.   The simulation continues to evaluation step (\ding{194}) where it is determined if desired state is achieved.  If it is achieved, the motor execution is enabled and the contents of loopy structure provide high-level context or working memory for plan execution.  If not, the simulation is reset with current need and continues. }
\end{figure}


What might be the evolutionary driving force that motivated System 2 thinking?
We conjecture that System 2 thinking is the byproduct of the solution to the navigational challenge in the dark cave, where mammals have to integrate paths for exploration and foraging.
The solution developed by mammals was to build a cognitive map of the exploration space using the hippocampus and related memory networks.  The hippocampus can be modeled as a loop-structured memory that enables repetition of a sequence to write down to cortical columns for episodic memory.  This structure is required because cortical columns in neocortex require repetition of a sequence to learn with Hebbian learning.  But the events in navigation or operant conditioning are not repetitive.   The hippocampus loop structure allows the repetition of previous events (replay) and the simulation of future events (preplay)~\cite{dragoi2011preplay}.  Replay and preplay are supported by world knowledge encoded in neocortex gray matter.

We model a  cortical column as an autoregressive memory that is given context signals from modules in higher layers and input signals from lower layers.  Those signals go through context and input poolers that reduce dimension and correct errors with latent inhibition.  The output of poolers are concatenated and fed into autoregressor that predicts the next signal given the current signal.  The poolers and autoregressors can be updated locally in self-supervised learning.  Therefore, the modules can be connected with multiple parents and children in heterarchical network.

  In the proposed cognitive architecture, the signals in the loop memory of the hippocampus become the top-most layer and connected to lower layers in cortical columns of the neocortex.    The predictions in the higher layer is used as context for lower-layers and activate the whole network in top-down ways. Lower-layers provide detailed sensory activation thus enabling richer simulation.  This simulation provides the mechanism for operant conditioning.  HICA explains how System 2 thinking  might be implemented in biological brain and provides a theory for building mechanical model for AI.

\bibliographystyle{plainnat}
\bibliography{HLAI}

\end{document}